\documentclass[a4paper]{llncs}
\usepackage{subfigure}
\usepackage{graphicx}
\usepackage{url}
\usepackage{multicol}
\usepackage{isolatin1}
\usepackage{xspace}
\usepackage{amsmath} 

\title{\mbox{Property analysis of symmetric travelling salesman} problem instances acquired through evolution}
\author{J.I. van Hemert}
\institute{Centre for Emergent Computing, Napier University, Edinburgh, UK\\\email{j.van.hemert@napier.ac.uk}}

\newcommand{\Ab}[1]
{%
{\scshape\lowercase{\mbox{#1}\xspace}}%
}

\begin{document}
\bibliographystyle{splncs}

\maketitle

\begin{abstract}
We show how an evolutionary algorithm can successfully be used to evolve a set of difficult to solve symmetric travelling salesman problem instances for two variants of the Lin-Kernighan algorithm. Then we analyse the instances in those sets to guide us towards deferring general knowledge about the efficiency of the two variants in relation to structural properties of the symmetric travelling salesman problem.
\end{abstract}

\section{Introduction}

The travelling salesman problem (\Ab{TSP}) is well known to be \Ab{NP}-complete. It is mostly studied in the form of an optimisation problem where the goal is to find the shortest Hamiltonian cycle in a given weighted graph \cite{LLRS1985}. Here we will restrict ourselves to the \emph{symmetric} travelling salesman problem, i.e., $\textrm{distance}(x,y) = \textrm{distance}(y,x)$, with Euclidean distances in a two-dimensional space. 

Over time, much study has been devoted to the development of better \Ab{TSP} solvers. Where ``better'' refers to algorithms being more efficient, more accurate, or both. It seems, while this development was in progress, most of the effort went into the construction of the algorithm, as opposed to studying the properties of travelling salesman problems. The work of \cite{CKT1991} forms an important counterexample, as it focuses on determining phase transition properties of, among others, \Ab{TSP} in random graphs, by observing both the graph connectivity and the standard deviation of the cost matrix. Their conjecture, which has become popular, is that all \Ab{NP}-complete problems have at least one order parameter and that hard to solve problem instances are clustered around a critical value of this order parameter.

It remains an open question whether the critical region of order parameters are mainly depending on the properties of the problem, or whether it is linked to the algorithm with which one attempts to solve the problem. However, a substantial number of empirical studies have shown that for many constraint satisfaction and constraint optimisation problems, a general region exists where problems are deemed more difficult to solve for a large selection of algorithms \cite{BPS2003,MZKST1999,Hayes1997}.

Often the characterisation of the order parameter includes structural properties \cite{Hogg1996,CG1999}, which leads to both a more accurate prediction and a better understanding of where hard to solve problems can be expected. Naturally, this does not exclude that a relationship between an algorithm and certain structural properties can exist. In this study, we shall provide empirical evidence for the existence of such a distinct relationship for two \Ab{TSP} problem solvers, which is of great influence on the efficiency of both algorithms.

In the following section we describe the process of evolving \Ab{TSP} instances. Then, in Section~\ref{s:lin_kernighan} we provide a brief overview of the Lin-Kernighan algorithm, and the variants used in this study. Section~\ref{s:experiments} contains the empirical investigation on the difficulty and properties of evolved problem instances. Last, in Section~\ref{s:conclusions} we provide conclusions.

\section{Evolving TSP instances}\label{s:evolving_tsp}
The general approach is similar to that in \cite{Hemert2003}, where an evolutionary algorithm was used to evolve difficult to solve binary constraint satisfaction problem instances for a backtracking algorithm. Here, we use a similar evolutionary algorithm to evolve difficult travelling salesman problem instances for two well known \Ab{TSP} solvers.

\begin{figure}
	\centering
	\vspace{-0.4\baselineskip}
	\includegraphics[width=0.68\textwidth]{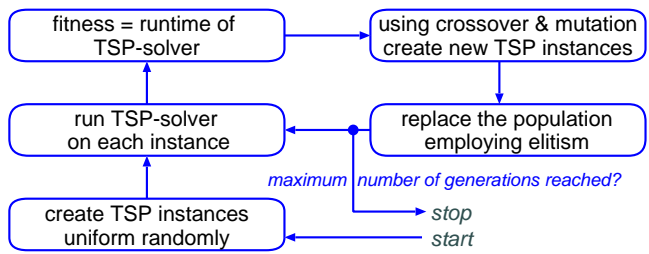}
	\caption{The process of evolving \Ab{TSP} instances that are difficult to solve}
	\label{fig:process}
	\vspace{-\baselineskip}
\end{figure}
A \Ab{TSP} instance is represented by a list of 100 $(x,y)$ coordinates on a $400\times 400$ grid. The list directly forms the chromosome representation with which the evolutionary algorithm works. For each of the 30 initial \Ab{TSP} instances, we create a list of 100 nodes, by uniform randomly selecting $(x,y)$ coordinates on the grid. This forms the first step in the process, depicted in Figure~\ref{fig:process}. Then the process enters the evolutionary loop: Each \Ab{TSP} instance is awarded a fitness equal to the search effort (defined in Section~\ref{s:lin_kernighan}) required by the \Ab{TSP} solver to find a near-optimal shortest tour. Using two-tournament selection, we repeatedly select two parents, which create one offspring using uniform crossover. Every offspring is subjected to mutation, which consists of replacing each one of its nodes with a probability $pm$, with uniform randomly chosen $(x,y)$ coordinates. This generational model is repeated 29 times, and together with the best individual from the current population (1-elitism), a new population is formed. The loop is repeated for 600 generations.

The mutation rate $pm$ is decreased over the subsequent generations. This process makes it possible to take large steps in the search space at the start, while keeping changes small at the end of the run. The mutation rate is varied using,
\begin{displaymath}
\textit{pm} = \textit{pm}_{\textit{end}} + (\textit{pm}_{\textit{start}} - \textit{pm}_{\textit{end}}) \cdot 2^{\frac{-\textit{generation}}{\textit{bias}}},
\end{displaymath}
from \cite{KLT2003} where the parameters are set as $\textit{bias} = 2$, $\textit{pm}_{\textit{start}} = 1/2$, $\textit{pm}_{\textit{end}} = 1/100$, and \emph{generation} is the current generation.

\section{Lin-Kernighan}\label{s:lin_kernighan}

As for other constrained optimisation problems, we distinguish between two types of algorithms, complete algorithms and incomplete algorithms. The first are often based on a form of branch-and-bound, while the latter are equipped with one or several heuristics. In general, as complete algorithms will quickly become useless when the size of the problem is increased, the development of \Ab{TSP} solvers has shifted towards heuristic methods. One of the most renowned heuristic methods is Lin-Kernighan \cite{LK1973}. Developed more than thirty years ago, it is still known for its success in efficiently finding near-optimal results.

The core of Lin-Kernighan, and its descending variants, consists of edge exchanges in a tour. It is precisely this procedure that consumes more than 98\% of the algorithm's run-time. Therefore, in order to measure the \emph{search effort} of Lin-Kernighan-based algorithms we count the number of times an edge exchange occurs during a run. Thus, this measure of the time complexity is independent of the hardware, compiler and programming language used. In this study, we use two variants of the Lin-Kernighan algorithm, which are explained next.

\subsection{Chained Lin-Kernighan}
\emph{Chained Lin-Kernighan} (\Ab{CLK}) is a variant \cite{ACR1999} that aims to introduce more robustness in the resulting tour by chaining multiple runs of the Lin-Kernighan algorithm. Each run starts with a perturbed version of the final tour of the previous run. The length of the chain depends on the number of nodes in the \Ab{TSP} problem.

In \cite{Papadimitriou1992}, a proof is given demonstrating that local optimisation algorithms that are \Ab{PLS}-complete (Polynomial Local Search), can always be forced into performing an exponential number of steps with respect to the input size of the problem. In \cite{JM1997}, Lin-Kernighan was first reported to have difficulty on certain problem instances, which had the common property of being clustered. The reported instances consisted of partial graphs and the bad performance was induced because the number of ``hops'' required to move the salesman between two clusters was set large enough to confuse the algorithm. We are using the symmetric \Ab{TSP} problem, where only full graphs exist and thus, every node can be reached from any other in one ''hop''.

\subsection{Lin-Kernighan with Cluster Compensation}
As a reaction on the bad performance reported in \cite{JM1997}, a new variant of Lin-Kernighan is proposed in \cite{Neto1999}, called \emph{Lin-Kernighan with Cluster Compensation} (\Ab{LK-CC}). This variant aims to reduce the computational effort, while maintaining the quality of solutions produced for both clustered and non-clustered instances.

Cluster compensation works by calculating the cluster distance for nodes, which is a quick pre-processing step. The cluster distance between node $v$ and $w$ equals the minimum bottleneck cost of any path between $v$ and $w$, where the bottleneck cost of a path is defined as the heaviest edge on that path. These values are then used in the guiding utility function of Lin-Kernighan to prune unfruitful regions, i.e., those involved with high bottlenecks, from the search space.

\section{Experiments}\label{s:experiments}

Each experiment consists of 190 independent runs with the evolutionary algorithm, each time producing the most difficult problem instance at the end of the run. With 29 new instances at each of the 600 generations, this results in running the Lin-Kernighan variant $3\,306\,000$ times for each experiment. The set of problem instances from an experiment is called \emph{Algorithm:Evolved set}, where \emph{Algorithm} is either \Ab{CLK} or \Ab{LK-CC}, depending on which problem solver was used in the experiment.

The total set of problem instances used as the initial populations for the 190 runs is called \emph{Random set}, and it contains $190 \times 30 = 5\,700$ unique problem instances, each of which is generated uniform randomly. This set of initial instances is the same for both Lin-Kernighan variants.

\subsection{Increase in difficulty}
\begin{figure}[ht]
	\centering
	\includegraphics[width=0.95\textwidth]{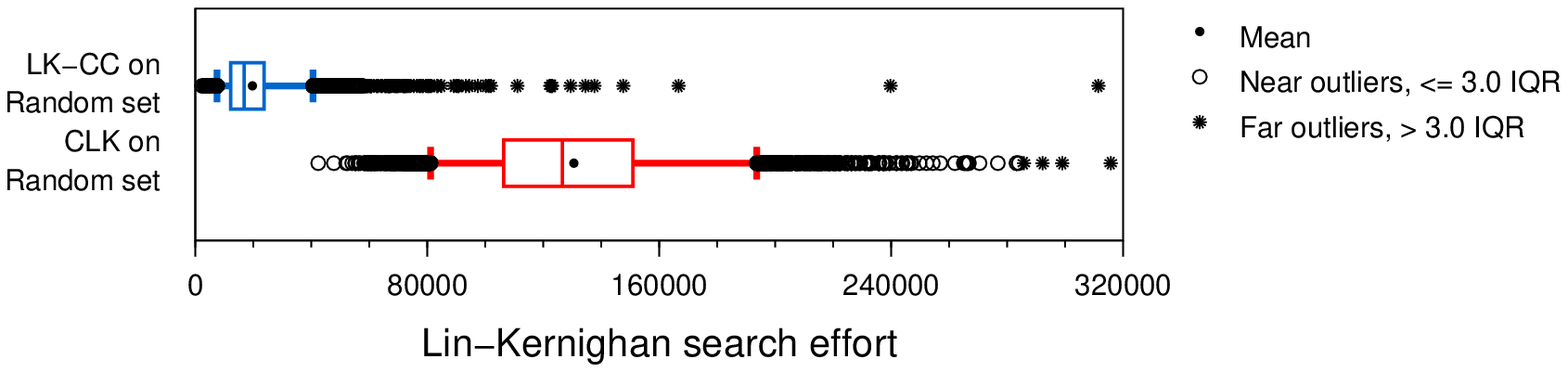}\\\vspace{5pt}
	\includegraphics[width=0.95\textwidth]{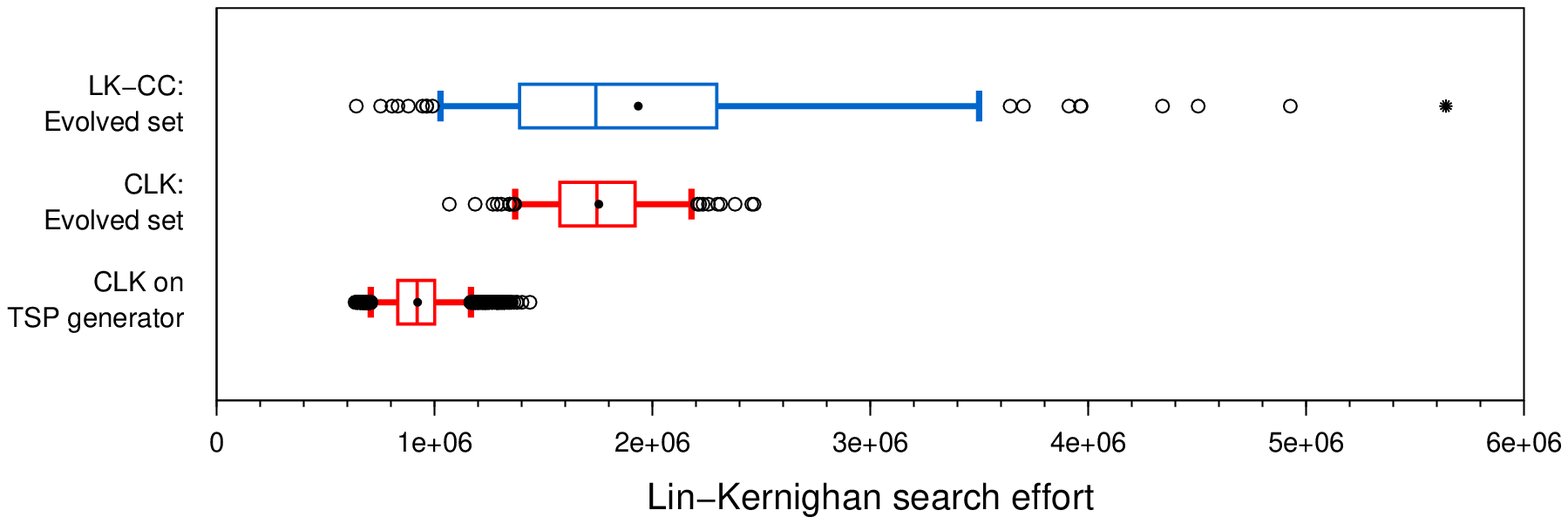}
	\caption{Box-and-whisker plots of the search effort required by \Ab{CLK} and \Ab{LK-CC} on the Random set (top), and \Ab{CLK} on the \Ab{TSP} generator and on the \Ab{CLK}:Evolved set (bottom) and by \Ab{LK-CC} on the \Ab{LK-CC}:Evolved set (bottom)}
	\label{fig:difficulty}
\end{figure}

In Figure~\ref{fig:difficulty}, we show the amount of search effort required by Chain Lin-Kernighan to solve the sets of \Ab{TSP} instances corresponding to the different experiments, as well as to the Random set. Also, we compared these results to results reported in \cite{HU2004}, where a specific \Ab{TSP} generator was used to create clustered instances and then solved using the Chained Lin-Kernighan variant. This set contains the 50 most difficult to solve instances from those experiments and it is called \emph{TSP generator}.

In Figure~\ref{fig:difficulty}, we notice that the mean and median difficulty of the instances in the \Ab{CLK}:Evolved set is higher than those created with the \Ab{TSP} generator. Also, as the 5/95 percentile ranges are not overlapping, we have a high confidence of the correctness of the difference in difficulty.

When comparing the difficulty of \Ab{CLK} and \Ab{LK-CC} for both the Random set and the Evolved sets in Figure~\ref{fig:difficulty}, we find a remarkable difference in the the amount of variation in the results of both algorithms. \Ab{CLK} has much more variation with the Random set than \Ab{LK-CC}. However, for the evolved sets, the opposite is true. We also mention that for the Random set, \Ab{LK-CC} is significantly faster than \Ab{CLK}, while difference in speed for the evolved sets is negligible.

\subsection{Discrepancy with the optimum}
We count the number of times the optimum was found by both algorithms for the Random set and for the corresponding Evolved sets. These optima are calculated using Concorde's \cite{ABCC1999} branch-and-cut approach to create an \Ab{LP}-representation of the \Ab{TSP} instance, which is then solved using Qsopt \cite{url_qsopt}. We show the average discrepancy between optimal tour length and the length of the tour produced by one of the problem solvers.

For the Random set, \Ab{CLK} has an average discrepancy of 0.004\% (stdev: 0.024), and it finds the best tour for 95.8\% of the set. For the same set of instances, \Ab{LK-CC} has an average discrepancy of 2.08\% (stdev: 1.419), and it finds the best tour for 6.26\% of the set.

A similar picture presents itself for the Evolved sets. Here, \Ab{CLK} has an average discrepancy of 0.03\% (stdev: 0.098), and find the best tour for 84.7\% of the \Ab{CLK}:Evolved set. \Ab{LK-CC} has an average discrepancy of 2.58\% (stdev: 1.666), and finds the best tour for 4.74\% of the \Ab{LK-CC}:Evolved set.

\subsection{Clustering properties of problem sets}
To get a quantifiable measure for the amount of clusters in \Ab{TSP} instances we use the clustering algorithm \Ab{GDBSCAN} \cite{SEKX1998}. This algorithm uses no stochastic process, assumes no shape of clusters, and works without a predefined number of clusters. This makes it an ideal candidate to cluster 2-dimensional spatial data, as the methods suffers the least amount of bias possible. It works by using an arbitrary neighbourhood function, which in this case is the minimum Euclidean distance. It determines clusters based on their density by first seeding clusters and then iteratively collecting objects that adhere to the neighbourhood function. The neighbourhood function here is a spatial index, which results in a run-time complexity of $O(n \log n)$.

\begin{figure}
	\centering
	\includegraphics[width=0.90\textwidth]{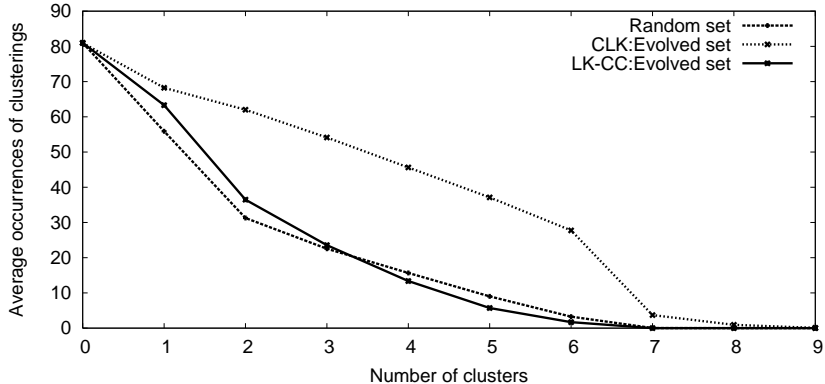}
	\caption{Average amount of clusters found for problem instances of the Random set and for problem instances evolved against \Ab{CLK} and \Ab{LK-CC}}
	\label{fig:gdbscan_clustering_properties}
\end{figure}

Clustering algorithms pose a large responsibility on their users, as every clustering algorithm depends on at least one parameter to help it define what a cluster is. Two common parameters are the number of clusters, e.g., for variants of $k$-means, and distance measurements to decide when two points are near enough to consider them part of the same cluster. The setting of either of these parameters greatly affects the resulting clustering. To get a more unbiased result on the number of clusters in a set of points we need a more robust method.

To get a more robust result for the number of clusters found in the different sets of \Ab{TSP} instances we repeat the following procedure for each instance in the set. Using the set $\{10,11,12,\ldots,80\}$ of settings for the minimum Euclidean distance parameter for \Ab{GDBSCAN}, we cluster the \Ab{TSP} instance for every parameter setting. We count the number of occurrences of each number of clusters found. Then we average these results over all the \Ab{TSP} instances in the set. The set of minimum Euclidean distance parameters is chosen such that it includes both the peak nd the smallest number of clusters for each problem instance.

We use the above procedure to quantify the clustering of problem instances in the Random set and the two evolved sets. Figure~\ref{fig:gdbscan_clustering_properties} shows that for the Random set and the \Ab{LK-CC}:Evolved set, the average number of clusters found does not differ by much. Instead, the problem instances in the \Ab{CLK}:Evolved set contain consistently more clusters. The largest difference is found for 2--6 clusters.

\subsection{Distribution of segment lengths}

\begin{figure}
	\centering
	\includegraphics[width=0.45\textwidth]{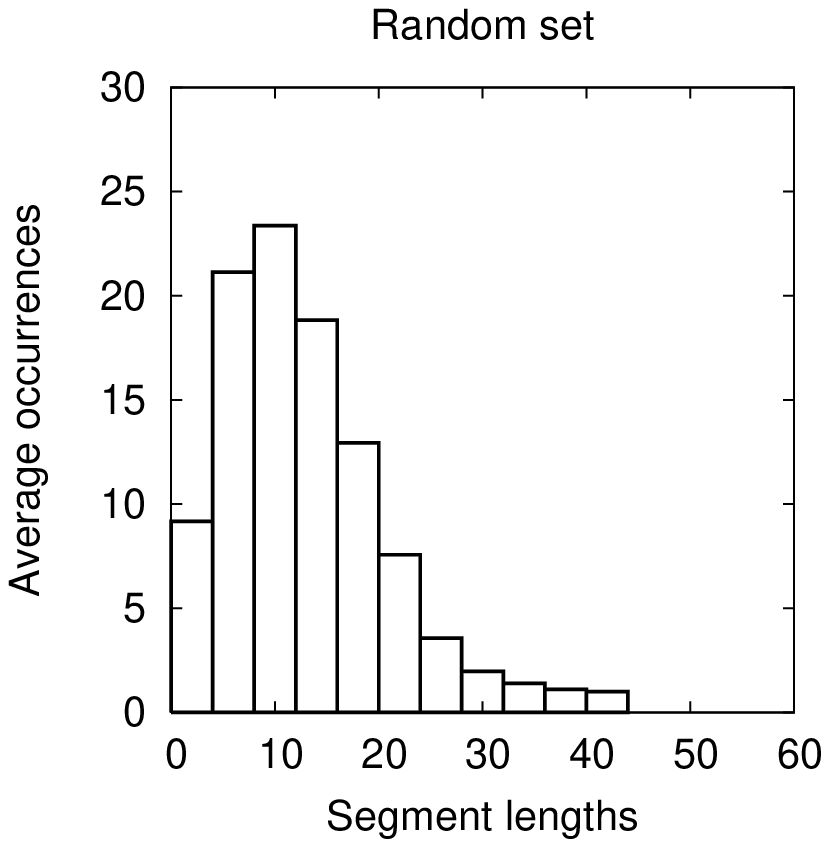}
	\includegraphics[width=0.45\textwidth]{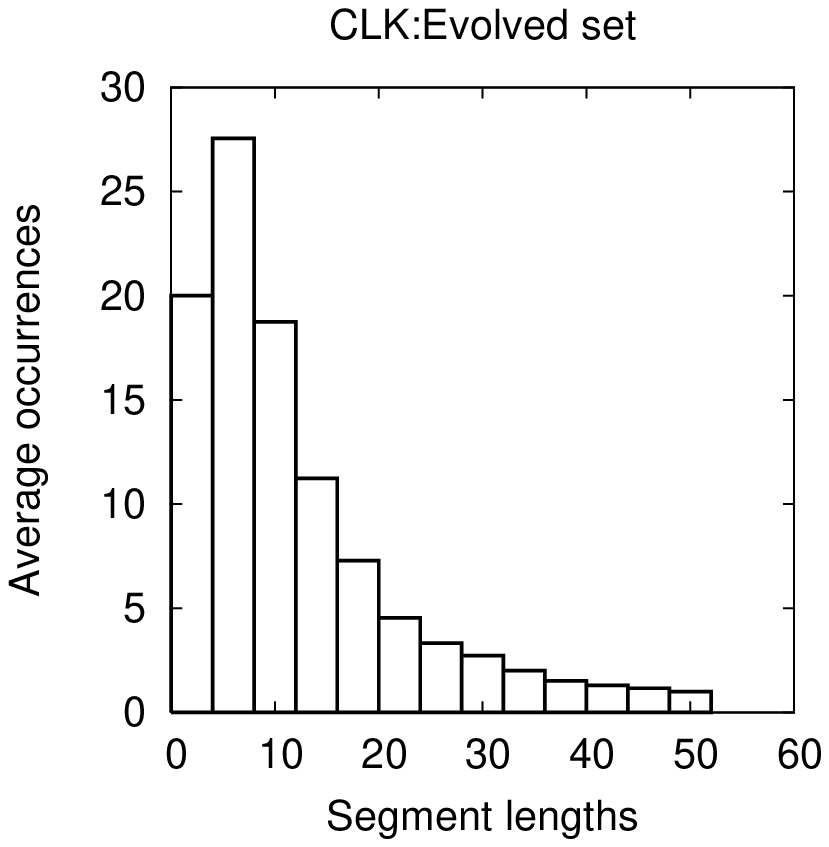}
	\caption{Average distribution of segment lengths in the resulting Chained Lin-Kernighan tour for \Ab{TSP} instances of the Random set and of the \Ab{CLK}:Evolved set}
	\label{fig:cumulative-histogram}
\end{figure}

For both problem solvers, we study the difference between the distribution of the segment lengths of resulting tours from both the Random set and the corresponding Evolved set. For both sets, we take the optimal tour of each \Ab{TSP} instance in the set and then, for each tour, observe all the segment lengths. Finally, we count the occurrences of the segment lengths, and average these over the whole set of tours.

Figure~\ref{fig:cumulative-histogram} shows the average distribution in segment lengths for tours derived from \Ab{TSP} instances from the Random set and for the \Ab{CLK}:Evolved set. We notice that the difference from the Random set to the \Ab{CLK}:Evolved set is the increase of very short segments (0--8) and  more longer segments (27--43), and the introduction of long segments (43--58). As the number of segments in a tour is always the same, i.e., 100, these increases relate to the decrease of medium length segments (9--26).

\begin{figure}
	\centering
	\includegraphics[width=0.90\textwidth]{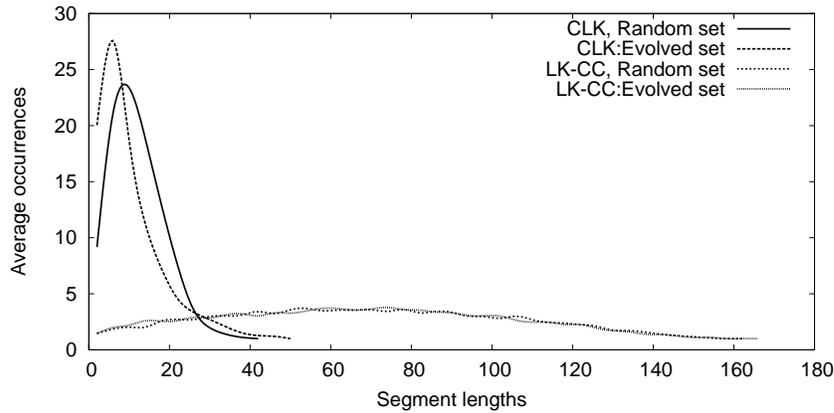}
	\caption{Average distribution of segment lengths in the resulting tours for both the Random set and Evolved set of problem instances for \Ab{CLK} and \Ab{LK-CC}}
	\label{fig:algorithm_histograms}
\end{figure}

Figure~\ref{fig:algorithm_histograms} shows the average distribution of segment lengths for both problem solvers. For \Ab{LK-CC}, we observe no significant changes in the distribution of segment lengths between the Random set and \Ab{LK-CC}:Evolved set.

When comparing the average distribution of segment lengths of tours in both problem solvers we clearly see a large difference. \Ab{LK-CC}, compared to \Ab{CLK}, uses much longer segments. Those segments most frequently used, in the range of 60--80, never occur at all with \Ab{CLK}. The distribution for \Ab{LK-CC} seems to match a flattened normal distribution, whereas the distribution for \Ab{CLK} is much more skewed, \Ab{CLK} favours the usage of short segments of a length less than 20.

\subsection{Distribution of pair-wise distances}

\begin{figure}
	\centering
	\includegraphics[width=0.90\textwidth]{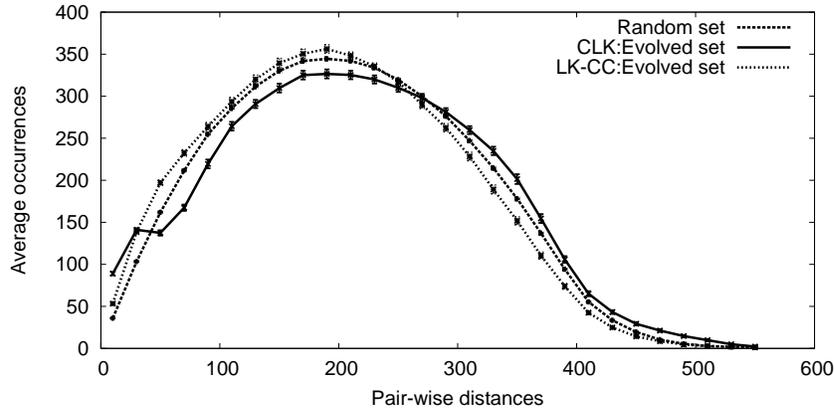}
	\caption{Distribution of distances over all pairs of nodes in randomly generated and evolved problem instances after 600 generations (\Ab{CLK} and \Ab{LK-CC}), 95\% confidence intervals included, most of which are small}
	\label{fig:tsp_histograms}
\end{figure}

In Figure~\ref{fig:tsp_histograms}, we show the average number of occurrences for distances between pairs of nodes. Every \Ab{TSP} instance contains $\binom{100}{2}$ pairs of nodes on the account that it forms a full graph. The distribution of these pair-wise distances mostly resembles a skewed Gaussian distribution. The main exception consists of the cut-off at short segments lengths. These very short distances, smaller than about 4, occur rarely when 100 nodes are distributed over a $400 \times 400$ space.

For the Chained Lin-Kernighan we notice a change in the distribution similar to that in the previous section. Compared with the Random set, both the number of short segments and the number of long segments increases. Although not to the same extent when observing the distribution of segment lengths. Also, the number of medium length occurrences is less than for the Random set. This forms more evidence for the introduction of clusters in the problem instances. Although this analysis does not provide us with the amount of clusters, it does give us an unbiased view on the existence of clusters, as it is both independent of the \Ab{TSP} algorithms and any clustering algorithm.

Also shown in Figure~\ref{fig:tsp_histograms} is the distribution of pair-wise distances for problem instances evolved against the \Ab{LK-CC} algorithm. While we notice an increase in shorter distances, this is matched by an equal decrease in longer distances. Thus, the total layout of the nodes becomes more huddled together.

\subsection{Swapping evolved sets}

We run each variant on the problem instances in the set evolved for the other variant. Table~\ref{tab:swapping_sets} clearly shows that a set evolved for one algorithm is much less difficult for the other algorithm. However, each variant needs significantly more search effort for the alternative Evolved set than for the Random set. This indicates that some properties of difficulty are shared between the algorithms.

\begin{table}[ht]
	\centering
	\caption{Mean and standard deviation, in brackets, of the search effort required by both algorithms on the Random set and both Evolved sets}
	\label{tab:swapping_sets}
\begin{tabular}{lrrrr}
	\hline\noalign{\smallskip}
	& \multicolumn{2}{c}{\Ab{CLK}} & \multicolumn{2}{c}{\Ab{CC-LK}} \\
	\noalign{\smallskip}\hline\noalign{\smallskip}
\Ab{CLK}:Evolved set & $1\,753\,790$ & $(251\,239)$ & $207\,822$ & $(155\,533)$\\
\Ab{CC-LK}:Evolved set & $268\,544$ & $(71\,796)$ & $1\,934\,790$ & $(799\,544)$ \\
Random set & $130\,539$ & $(34\,452)$ & $19\,660$ & $(12\,944)$ \\
	\noalign{\smallskip}\hline
\end{tabular}
\end{table}

\section{Conclusions}\label{s:conclusions}

We have introduced an evolutionary algorithm for evolving difficult to solve travelling salesman problem instances. The method was used to create a set of problem instances for two well known variants of the Lin-Kernighan heuristic. These sets provided far more difficult problem instances than problem instances generated uniform randomly. Moreover, for the Chained Lin-Kernighan variant, the problem instances are significantly more difficult than those created with a specialised \Ab{TSP} generator. Through analysis of the sets of evolved problem instances we show that these instances adhere to structural properties that directly afflict on the weak points of the corresponding algorithm. 

Problem instances difficult for Chained Lin-Kernighan seem to contain clusters. When comparing with the instances of the \Ab{TSP} generator, these contained on average more clusters (10 clusters) then the evolved ones (2--6 clusters). Thus, this leads us to the conjecture that clusters on itself is not sufficient property to induce difficulty for \Ab{CLK}. The position of the clusters and distribution of cities over clusters, as well as the distribution of nodes not belonging to clusters, can be of much influence.

The goal of the author of \Ab{LK-CC} is to provide a \Ab{TSP} solver where its efficiency and effectiveness are not influenced by the structure of the problem \cite{Neto1999}. Problem instances evolved in our study, which are difficult to solve for Lin-Kernighan with Cluster Compensation, tend to be condense and contain random layouts. The  algorithm suffers from high variation in the amount of search effort required, therefore depending heavily on a lucky setting of the random seed. Furthermore, its effectiveness is much lower than that of \Ab{CLK}, as the length of its tours are on average, further away from the optimum. Thus, it seems that to live up its goals, \Ab{LK-CC} is losing on both performance and robustness.

The methodology described here is of a general nature and can, in theory, be used to automatically identify difficult problem instances, or instances that inhibit other properties. Afterwards, these instances may reveal properties that can lead to general conclusions on when and why algorithms show a particular performance. This kind of knowledge is of importance when one needs to select an algorithm to solve a problem of which such properties can be measured.

\section*{Acknowledgements}
The author is supported through a TALENT-Stipendium awarded by the Netherlands Organization for Scientific Research (NWO).

\bibliography{paper}
\end{document}